\documentclass[lettersize,journal]{IEEEtran}
\usepackage{amsmath,amsfonts}
\usepackage[ruled,linesnumbered]{algorithm2e}
\usepackage{array}
\usepackage{textcomp}
\usepackage{stfloats}
\usepackage{url}
\usepackage{verbatim}
\usepackage{graphicx}
\usepackage{cite}
\usepackage{verbatim}
\usepackage{graphicx}
\usepackage{textcomp}
\usepackage{booktabs} 
\usepackage{setspace}
\usepackage{subfigure}
\usepackage{subfiles}  
\usepackage{caption}
\usepackage{mathrsfs}
\usepackage{multirow}
\usepackage{xcolor}
\usepackage[numbers,sort&compress]{natbib}
\newcommand{\name}{SAFE-D~}
\newcommand{\sname}{SAFE-D}

\begin{document}
\title{SAFE-D: A Spatiotemporal Detection Framework for Abnormal Driving Among Parkinson's Disease-like Drivers}

\author{Hangcheng Cao,~
        Baixiang Huang,~
        Longzhi Yuan,~
        Haonan An,~
        Zihan Fang,~
        Xianhao Chen,~
        and Yuguang Fang,~\IEEEmembership{Fellow,~IEEE}

\thanks{The research work described in this paper was conducted in the JC STEM Lab of Smart City funded by The Hong Kong Jockey Club Charities Trust under Contract 2023-0108. This work was also supported in part by the Hong Kong SAR Government under the Global STEM Professorship and Research Talent Hub. The work of X. Chen was supported in part by the Research Grants Council of Hong Kong under Grant 27213824.}

\thanks{Hangcheng Cao, Longzhi Yuan, Haonan An, Zihan Fang,  and Yuguang Fang are with Hong Kong JC STEM Lab of Smart City and Department of Computer Science, City University of Hong Kong,
Hong Kong, China (e-mail: hangccao@cityu.edu.hk; longyuan@cityu.edu.hk; haonanan2-c@my.cityu.edu.hk; zihanfang3@cityu.edu.hk; my.fang@cityu.edu.hk) 

Baixiang Huang is with the Department of Computer Science and Engineering, Hong Kong University of Science and Technology, Hong Kong, China. (e-mail: xiaoyuanzi22333@gmail.com) 

Xianhao Chen is with the Department of Electrical and Electronic Engineering, The University of Hong Kong, Hong Kong, China (e-mail: xchen@eee.hku.hk).
}

}      

\maketitle

\begin{abstract}
A driver's health state serves as a determinant factor in driving behavioral regulation. Subtle deviations from normalcy can lead to operational anomalies, posing risks to public transportation safety. While prior efforts have developed detection mechanisms for \textit{functionally-driven temporary anomalies} such as drowsiness and distraction, limited research has addressed \textit{pathologically-triggered deviations}, especially those stemming from \textit{chronic medical conditions}. To bridge this gap, we investigate the driving behavior of Parkinson’s disease patients and propose \sname, a novel framework for detecting Parkinson-related behavioral anomalies to enhance driving safety. Our methodology starts by performing analysis of Parkinson's disease symptomatology, focusing on primary motor impairments, and establishes causal links to degraded driving performance. To represent the subclinical behavioral variations of early-stage Parkinson’s disease, our framework integrates data from multiple vehicle control components to build a behavioral profile. We then design an attention-based network that adaptively prioritizes spatiotemporal features, enabling robust anomaly detection under physiological variability. Finally, we validate \name on the Logitech G29 platform and CARLA simulator, using data from three road maps to emulate real-world driving. Our results show \name achieves 96.8\% average accuracy in distinguishing normal and Parkinson-affected driving patterns.
\end{abstract}


\section{Introduction}
\label{sec:intro}
The health states of a driver exert a deterministic influence on driving behaviors, serving as a pivotal determinant of road safety. Behavioral anomalies that arise from health state deviations increase the potential risk of a traffic accident occurrence, thus compromising driving safety. Current research primarily targets at the detection of abnormal driving behaviors linked to temporary functional impairments, notably drowsiness and external distractions. For example, in-vehicle monitoring systems track behaviors such as blink frequency~\cite{baccomera,shahneous} and head pose dynamics~\cite{yanigueview,subasi2022eeg,xie2020real} to assess these transient states. In contrast, behavioral deviations originating from chronic pathological states remain underexplored. To address this issue, our study investigates the influence of Parkinson's disease (PD)~\cite{bloemparkinson} on driving safety, a prevalent clinically defined neurodegenerative disorder characterized by motor symptoms such as tremors and bradykinesia. Notably, prior research has reported that approximately 82\% of individuals with PD hold a valid driver's license, with 60\% actively engaged in driving~\cite{crizzkinson}. Therefore, our study aims to elucidate how PD-driven motor dysfunction manifests, thereby establishing detection frameworks for such pathology-specific abnormal driving behaviors. 

Current PD diagnostic research focuses mainly on identifying pathological markers in static clinical settings, such as computed tomography scans \cite{milekovspinal}, pharmacological challenges~\cite{aarslankinson}, and neurophysiological evaluations~\cite{schestysiologic}. While these methods provide clinical validity, their reliance on professional facility-based assessments limits continuous disease monitoring. Recent advances in home-based ambulatory sensing address this gap through approaches such as smartphone interaction tremor analysis~\cite{kuosmanen2020let,zhang2022simple,ling2024model}, gait kinematics through trajectory imaging~\cite{zhang2020pdlens,jovaltimodal}, and nocturnal respiratory pattern tracking~\cite{yangtificial}. However, both clinical and home-based paradigms remain confined to static environments, failing to capture PD’s behavioral manifestations in dynamic driving contexts. Moreover, although preliminary medical studies~\cite{anjemark2023car,brock2022driving,gotardi2022parkinson} have discussed the physiological impact of PD on drivers, they lack granular analysis of symptom-behavior causality, specifically, how cardinal PD motor deficits translate to vehicular control anomalies. Therefore, systematic research on the driving behaviors of PD patients remain relatively limited. This critical knowledge gap underscores the necessity for investigations linking PD pathophysiology to operationally defined driving behaviors.

\begin{figure}[t]
\centering
\includegraphics[width=0.4\textwidth]{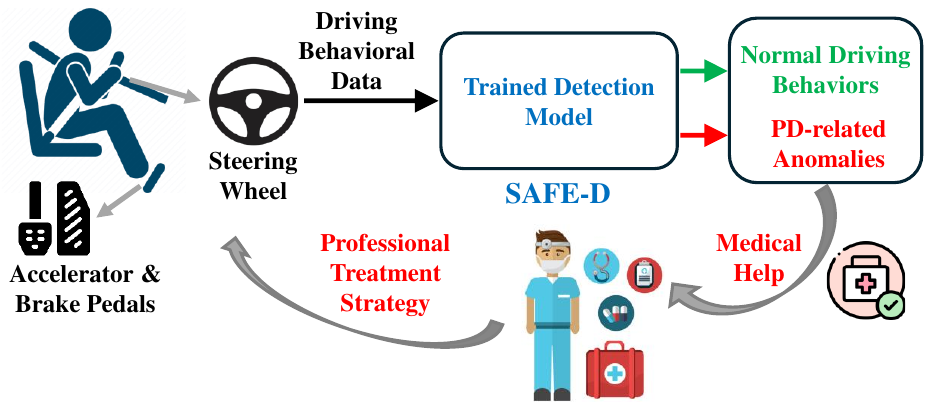}
\caption{Application scenario of our detection framework: leveraging sensing data from in-vehicle control components to identify abnormal driving behaviors and develop a detection model.}
\label{fig:workMechanism}
\end{figure}

Given the limitations of existing approaches, we present \sname, a novel framework designed to detect PD-related abnormal behaviors in driving environments, with architectural details illustrated in Fig.~\ref{fig:workMechanism}. The framework integrates the sensing data from multiple vehicle-mounted sensors to characterize PD-related behavioral deviations, thereby elucidating the mechanistic link between neurodegeneration and driving performance degradation. Specifically, PD-induced neurodegeneration compromises patients’ motor functions (e.g., tremors and bradykinesia), which are manifested as anomalies in vehicular control. Such behavioral signatures are captured through existing driver-vehicle interfaces, including steering wheel torque sensors and pedal pressure transducers. The acquired driving behavior data is processed by our detection model to identify PD-correlated operational patterns. Critically, \name requires no auxiliary hardware, enabling plug-and-play deployment through standard vehicle architectures. This design offers open access for continuous PD monitoring, allowing drivers to proactively assess neurological risks during routine commutes. When potential PD-related anomalies are detected, users receive contextualized behavioral insights to inform clinical decision-making\footnote{As PD diagnosis necessitates multidimensional evaluation~\cite{mostafa2019examining,milekovspinal,aarslankinson} such as neurological examinations and biomarker analysis, \name aims to provide complementary behavioral evidence to enhance diagnostic accuracy. The framework serves as a bridge between clinical assessments and driving performance detection, offering actionable items for personalized healthcare.}.
Although our proposed framework demonstrates potential for advancing PD and related behavior detection, significant design challenges are still ahead of us: i) \textit{Which PD symptoms directly affect what driving behaviors?} PD manifests through heterogeneous neuromotor pathologies, each differentially impacting driving performance. Prioritizing symptoms with direct vehicular control implications establishes causality relationship to guide our analysis; ii) \textit{What are the specific effects on driving behaviors?} Translating clinical symptoms into quantifiable driving behaviors, e.g., steering angular deviation variance and acceleration patterns, enables extraction of pathology-sensitive behavioral signatures; iii) \textit{How can the behavioral variations induced by PD be represented accurately?} Characterizing the driving behavioral discrepancies between PD individuals and healthy drivers is pivotal in both data analysis and the development of detection models. By systematically comparing these differences, it becomes possible to extract relevant features for \sname, thereby ensuring accurate detection.

To address these research questions, we develop the following methodology: i) Leveraging systematic evidence synthesis from clinical neuroscience literature~\cite{ling2024model,kuosmanen2020let,jovaltimodal,bloemparkinson,anjemark2023car,UPDRS}, we decompose PD manifestations into motor and non-motor symptoms. Motor deficits, including impaired coordination and movement execution, directly affect a driver’s fine motor control. In contrast, non-motor symptoms primarily influence quality of life and indirectly affect driving. Through analysis and interviews with movement disorder specialists, we identify three key motor symptoms (i.e., tremors, rigidity, and bradykinesia) as most impactful on driving performance.
ii) Based on motor symptom analysis, we link these deficits to specific impairments in driving, focusing on three core control components: steering, acceleration, and braking. In straight-line driving, tremors can hinder steady steering control, affecting lane-keeping. In non-straight-line scenarios, greater coordination among all controls is required; bradykinesia and rigidity may impair smooth, synchronized actions, reducing driving stability. This decomposition enables a detection framework connecting PD motor symptoms to observable driving anomalies. 
iii) After collecting road segment data, we apply a spatiotemporal fusion strategy with attention mechanisms to capture PD-related motor effects on driving. The dataset includes sensor readings from steering and pedal operations, offering insights into driver-vehicle interaction. To integrate heterogeneous sources and detect anomalies, we use an attention-based model that dynamically weights global and local spatiotemporal features, emphasizing motor control abnormalities and linking PD pathophysiology to abnormal driving.
Our study finally integrates the Logitech G29 driving system with the CARLA simulator to collect driver behavioral data\footnote{This configuration facilitates high-fidelity driving data collection, which has been widely employed in previous research on driver distraction~\cite{3652181,3581335,10265761} and drowsiness~\cite{9507391,10382460}}. The dataset encompasses three representative daily driving scenarios: urban, rural, and mixed environments. Experimental validation demonstrates that our proposed framework achieves an average accuracy of 96.8\% in distinguishing between normal and abnormal driving behaviors.

In summary, our study makes the following key contributions.
\begin{itemize}
    \item Within dynamic driving environments, we systematically analyze PD's impact on driver behaviors. We propose a framework for detecting abnormal behaviors, aiming to precisely identify anomalies attributed to the disease.

    \item We perform an extensive examination of PD motor symptoms, emphasizing their direct influence on driving behaviors. This analysis focuses on the specific manifestations and dynamic variations of these symptoms across both straight-line and non-straight-line driving scenarios.

    \item We integrate data from three in-vehicle sensors to represent driving behaviors. By employing attention mechanisms, we fuse spatiotemporal information to enhance the model’s sensitivity to abnormal behaviors.

    \item We collect data using widely adopted driving platforms and rigorously evaluate our proposed framework. Experimental findings indicate that \name provides significantly high accuracy in recognizing abnormal behaviors.
\end{itemize}

\section{Related Work}
\label{sec:relatedWork}
In this section, we provide a review of closely related work.

\subsection{Abnormal Driving Behavior Monitoring}
Driving safety has emerged as a critical research frontier, particularly in abnormal driving behavior monitoring. Current investigations predominantly focus on two prevalent functional anomalies: drowsiness~\cite{baccomera} and distraction~\cite{9758635}, while emerging studies are expanding into complex behavioral activities like affective state dynamics~\cite{3460938,8859275}. The new developed methodologies typically employ computer vision techniques to analyze physiological biomarkers including eye movement~\cite{7478592} and head posture patterns~\cite{9507391} for abnormal state assessment. For example, during drowsiness episodes, drivers exhibit quantifiable biomarkers such as increased blink rate, prolonged eyelid closure duration, and cephalic tilt angle deviations, which serve as reliable drowsiness indicators. However, unimodal approaches remain constrained by environmental interference factors, such as ambient illumination variance and occlusions. This limitation has driven the adoption of heterogeneous sensor fusion frameworks~\cite{519267,9507390} that synergize visual, auditory, and vehicular telemetry data to develop robust behavioral analytics models. Despite methodological advancements, contemporary research mostly addresses functional impairments, with limited attention to pathological mechanisms underlying driving anomalies. For example, neurological disorders (e.g., Parkinsonian syndromes) can induce catastrophic behavioral deviations, such as loss of vehicular control, that conventional functional abnormality models cannot address.

\subsection{Parkinson's Disease-related Anomaly Detection}
Addressing the research gap in driver pathological state analysis, this study explores the impact of PD-induced abnormalities on vehicular control. As a progressive neurodegenerative disorder, PD displays cardinal motor and non-motor symptoms that significantly impair daily activities and pose substantial risks to driving safety. Current PD-induced pathological changes and abnormal states primarily rely on static detection paradigms. A comprehensive evaluation protocol, which includes neurological examinations, neuroimaging, and standardized motor assessments, serves as the foundation for detecting pathological abnormalities associated with PD and guides therapeutic interventions such as pharmacotherapy and neuromodulation techniques~\cite{aarslankinson,schestysiologic}. Complementing clinical practice, smart health monitoring systems employ wearable sensors~\cite{ling2024model,3300313,9995237,3659627} and contactless sensing technologies~\cite{jovaltimodal} to continuously track abnormal motor fluctuations, autonomic responses, and medication efficacy, thereby detecting PD-induced abnormal behaviors for personalized care. However, these detection approaches exhibit limited effectiveness in dynamic driving contexts, where environmental variability interacts with PD pathophysiology. Although notable detection precision for PD-induced abnormalities has been achieved in clinical settings, there remains a significant need for detection frameworks that integrate clinical insights with driving dynamics. \name bridges this gap by transferring validated PD-induced abnormality detection methodologies to comprehensive pathological behavior monitoring in driving conditions.

\section{Background and Preliminaries}
\label{sec:background}
In this section, we present an analysis of PD symptoms and their impacts on driving behaviors, followed by the rational assumptions that underpin our study.

\subsection{Symptoms of Parkinson's Disease}
PD is a prevalent neurodegenerative disorder characterized by the progressive degeneration of dopaminergic neurons in the substantia nigra, leading to dopamine depletion and resulting in motor dysfunction~\cite{bloemparkinson}. Clinically, PD manifests through motor impairments such as tremors, bradykinesia, rigidity, postural instability, and gait abnormalities, alongside non-motor comorbidities, including cognitive \& psychiatric issues, sleep disorders, and autonomic dysfunction. We present the details of these symptoms in TABLE~\ref{table1}. Supported by empirical evidence~\cite{anjemark2023car, brock2022driving, gotardi2022parkinson, UPDRS}, our study focuses on tremors, bradykinesia, and rigidity, as these symptoms most directly impact driving performance. Tremors typically manifest as muscle shaking when gripping the steering wheel, significantly destabilizing steering control and vehicle handling precision. Bradykinesia leads to delayed reaction times, while rigidity restricts the range of motion, particularly when operating the steering wheel and pedals, resulting in instability in vehicle control. By examining these three primary motor symptoms, our research aims to explore how their manifestations can be used to identify abnormal driving behaviors.
\begin{table*}[t]
\caption{Basic information of Parkinson's symptoms and its impact on driving behaviors}
{\footnotesize
\label{table1}
\centering
\begin{tabular}{p{1.5cm}|p{3.5cm}|p{7.5cm}|p{2cm}}
\toprule
Type & Name & Symptom & Impact on Driving\\
\midrule
\multirow{9}{*}{Motor} & Tremor & Affecting the fingers or legs, characterized by continuous tremor, especially during resting periods & Direct\\
& Rigidity & Increased muscle tone leading to stiffness in the limbs and trunk, restricting movement& Direct\\
& Bradykinesia & Slowed movements and difficulty initiating actions, making daily activities harder to perform & Direct\\
& Postural Instability & Reduced balance, leading to frequent falls, particularly when walking or changing positions & Indirect \\
& Gait Abnormalities & Short, shuffling steps with reduced arm swinging, often resulting in a dragging gait & Indirect\\
\midrule
\multirow{7}{*}{Non-motor} & Cognitive \& Psychiatric Issues & Declining memory, difficulty concentrating, and impaired decision-making, which may progress to dementia & Indirect\\
& Sleep Disorders & Includes insomnia, frequent awakenings at night, sleep behavior disorder, and excessive daytime sleepiness & Indirect\\
& Autonomic Dysfunction & Issues like low blood pressure, constipation, frequent urination, and excessive salivation & Indirect\\
\bottomrule
\end{tabular}
}
\end{table*}
\subsection{The Impact of Motor Symptoms}
\label{subsc:sympotm}
\begin{figure}[t] 
\centering
\subfigure[Straight-line]{
\begin{minipage}[t]{0.48\linewidth}
\centering
\includegraphics[width=1\textwidth]{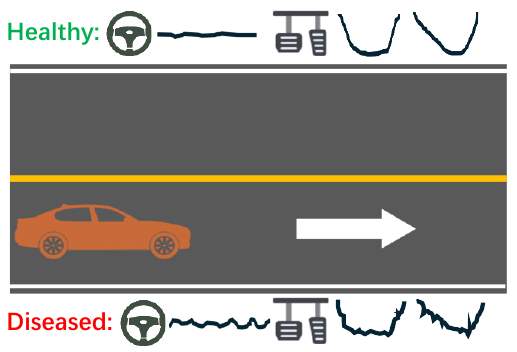}
\label{fig:line1}
\end{minipage}
}
\subfigure[Non-straight-line]{
\begin{minipage}[t]{0.38\linewidth}
\centering
\includegraphics[width=1\textwidth]{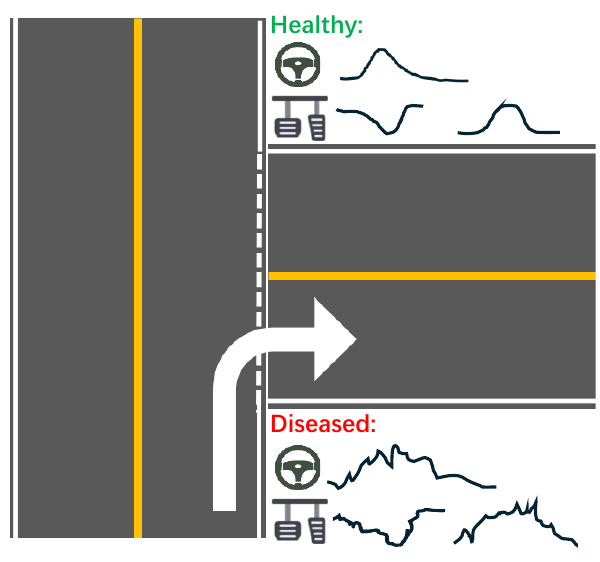}
\label{fig:nonline1}
\end{minipage}
}
\caption{Illustration of driving behavior difference between health and diseased drivers in (a) straight-line and (b) non-straight-line driving cases.}
\label{fig:lineDisease1}
\end{figure}

In real-world driving environments, road conditions are inherently diverse, leading to corresponding variations in driving behaviors that PD may manifest. To facilitate analysis and identification of abnormal behaviors, we classify road segments into two fundamental categories: straight-line and non-straight-line segments. For the former one, the primary driving task is to maintain the vehicle's linear trajectory with minimal steering adjustments. However, in this relatively static driving condition, the tremor symptoms of PD patients become particularly pronounced. Tremors, typically occurring at a frequency of 4 to 6 Hz~\cite{UPDRS}, introduce involuntary muscle movements that directly affect steering control. As a result, these tremors cause instability in sensory data, manifesting as \textit{continuous fluctuations} in wheel and pedal control behaviors, as illustrated in Fig.~\ref{fig:line1}. In contrast, non-straight road segments, such as right turns and other complex maneuvers, require precise and coordinated control of three control components. For drivers with PD, the presence of rigidity and bradykinesia significantly impairs their ability to execute these movements stably. Rigidity restricts motion flexibility, while bradykinesia slows down response times, leading to an overall reduction in stable control. Consequently, these impairments result in instability across three key sensory data streams, manifesting as \textit{sudden variations} in driving actions. For example, during a right turn, affected drivers often exhibit a lack of smooth continuity in their movements, leading to abrupt changes in steering and pedal control. These sudden variations shown in Fig.~\ref{fig:nonline1}, reflect the driver’s struggle to maintain fluid control over the vehicle. While PD can contribute to a wide range of abnormal driving behaviors, this study primarily characterizes these abnormalities through two key indicators: continuous fluctuations during straight-line driving and sudden variations during non-straight driving. These features directly capture the impact of PD-related motor impairments on driving operations.

\subsection{Assumptions}
To construct our detection framework, we establish foundational assumptions consistent with established methodologies in driver behavior analysis~\cite{azadani2021driving,baccomera,shahneous,subasi2022eeg,xie2020real,juncen2023mmdrive} and transport traffic research~\cite{fang2021spatial,wang2021prediction,luo2022estnet}:
\begin{itemize}
    \item {Drivers are instructed to maintain their natural driving patterns during the experimental trials, avoiding intentional modifications that could introduce measurement artifacts. This protocol ensures data fidelity by minimizing contamination from volitional influences, thereby preserving the authenticity of the captured driving patterns.}
    \item {Participants are presumed to be experienced drivers with sufficient proficiency to sustain stable and consistent driving behaviors. Novice drivers, due to their lack of expertise, may exhibit irregularities that resemble disease-induced anomalies, potentially confounding the analysis.}
    \item {Based on traffic flow information, we can obtain the natural driving data of the driver and exclude abnormal data segments caused by external factors such as vehicle breakdown and accidents.}
\end{itemize}
In summary, given the progressive nature of PD, the detection mechanism is assumed to operate through continuous monitoring. Longitudinal data analysis, supplemented by expert evaluation, facilitates the differentiation between pathological abnormalities and variations arising from unfamiliar environments or external influences.

\section{System Design}
\label{sec:sysDesign}

\subsection{System Overview}
\begin{figure}[t]
\centering
\includegraphics[width=0.44\textwidth]{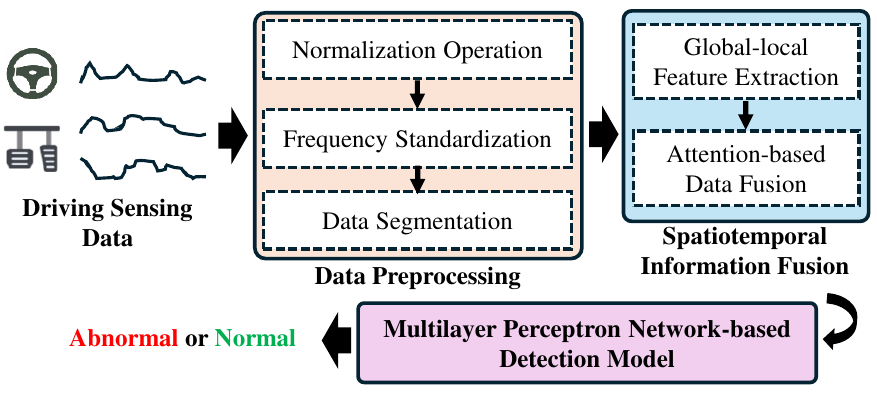}
\caption{Design of \sname, including three main modules: data preprocessing, information fusion, and detection model.}
\label{fig:safedFramework}
\end{figure}

We present the design of \name in Fig.~\ref{fig:safedFramework}, which leverages data from in-vehicle sensors to identify PD-related abnormal driving behaviors. The process begins with data preprocessing, where we standardize raw multi-channel inputs to maintain a uniform format, ensuring smooth integration into the subsequent network pipeline. During the information fusion phase, we incorporate spatiotemporal data to capture both global and local features, effectively representing individual driving actions. To enhance feature integration, we utilize multi-attention modules, enabling channel-aware and frame-aware fusion of relevant behavior information. Finally, we implement a detection module based on a multilayer perceptron network, harnessing the fused features to recognize abnormal samples.

\subsection{Data Preprocessing}
\label{subsec:dataProcessing}
This study introduces a preprocessing pipeline for time-series data collected from three channels: \textit{the steering wheel, accelerator pedal, and brake pedal}. To enhance the stability of our detection model training process, we first apply tailored data normalization strategies according to the physical characteristics of each channel. Specifically, the steering wheel angle range is transformed to [-1, 1], while the pedals travel percentages ranging from 0-100\% are normalized to [0, 1]. Following this, we address the issue of synchronizing multi-source data with differing sampling frequencies, we employ the nearest neighbor interpolation to align the data in the time domain. This operation makes the sampling rate of all channels to 30~\!Hz. In the segmentation phase, we apply fixed-length time windows to segment various driving scenarios (e.g., straight driving and turning), and a base analysis window of 4 seconds with a 1-second overlap. The effectiveness of this configuration is validated through our experiments, as presented in Section~\ref{sec:evaluation}. 

\begin{figure*}[b]
\centering
\includegraphics[width=0.88\textwidth]{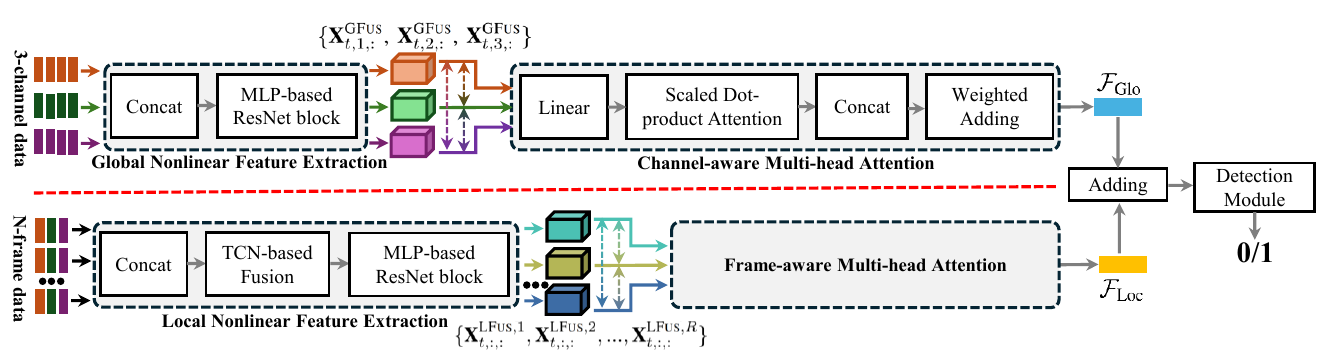}
\caption{Attention-based local and global spatiotemporal feature fusion network for driving behavior detection.}
\label{fig:architecture}
\end{figure*}
\subsection{Spatiotemporal Information Fusion}
\label{subsec:feature}
\textbf{The definition of detection model.} Driver behavior detection using in-vehicle sensor data is fundamentally a classification task of multi-dimensional time series. In our study, the preprocessed data is denoted as $\textbf{X} = \{{X_1},...,{X_T}\}$, where $T$ represents the total number of segmented samples. Each ${X_t} \in \mathbb{R}^{C \times M}$ captures the data for a specific driving action, with $C$ being the number of data channels and $M$ the number of ordered elements in each channel. Given the detection model $f_\theta(\cdot)$, our objective is to minimize a loss function as described as follows:
\begin{equation}
\label{eq:definition}
\min \ell(f_\theta({X_t}), y_t)
\end{equation}
In this context, the function $\ell(\cdot)$ represents the cross-entropy loss, quantifying the discrepancy between predicted and actual behavior labels. Subsequently, we delineate the architecture and computing methodology of the model $f_\theta(\cdot)$. The tri-channel data encompasses substantial spatiotemporal information, each channel manifesting the evolving driving dynamics over time. Moreover, individual data samples encapsulate complete driving actions (e.g., turning left and lane change), while more granular data frames reveal detailed local anomalies. Therefore, our framework concurrently leverages both global and local information, enhancing the model's ability to identify deviations in driving patterns at two distinct levels.

\begin{figure}[t]
\centering
\includegraphics[width=0.5\textwidth]{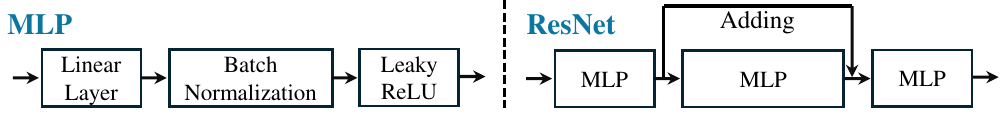}
\caption{Architecture of ResNet blocks.}
\label{fig:desenet}
\end{figure}

\textbf{Global feature extraction.}
The architecture of the two-type feature extraction network is illustrated in Fig.~\ref{fig:architecture}. To extract global features from a single-sample input, ${X_t}$ (i.e., $\textbf{X}_{t,:,:}$) is first passed through a non-linear feature extraction module. During this process, data from all three channels are concatenated, allowing subsequent (ResNet)~\cite{he2016deep} blocks to automatically capture interaction relationships across different channels and further refine non-linear information extraction. Using skip connections, the ResNet module effectively alleviates the problem of vanishing gradients in deep networks, ensuring smooth propagation of information across multiple layers. As illustrated in Fig.~\ref{fig:desenet}, the ResNet architecture comprises three multilayer perceptron (MLP)~\cite{cao2022handkey} units. Each MLP incorporates a linear layer, a BatchNorm layer, and a Leaky ReLU activation function, working synergistically to extract spatiotemporal features in a progressive manner. The final output of this module is three-feature matrices (corresponding to three channels) with the size of $64 \times 90$, denoted as $\{\textbf{X}_{t,{1},:}^{\textsc{GFus}}$, $\textbf{X}_{t,{2},:}^{\textsc{GFus}}$, $\textbf{X}_{t,{3},:}^{\textsc{GFus}}\}$. Subsequently, these matrices are fed to the channel-aware multihead attention (MHA)~\cite{zhaoethinking} module, where deeper feature refinement and enhanced cross-channel interactions occur. In the process of utilizing MHA for feature fusion, we begin by performing pairwise interactions between the matrices of the three channels, then compute the correlation between different channels and dynamically adjust their respective weights to facilitate efficient feature integration. For any selected two matrices (making the first two as an example) originating from distinct channels, they initially projected into query (Q), key (K), and value (V) subspaces through the following transformations:
\begin{equation}
\begin{aligned}
Q_1 &= \textbf{X}^{\textsc{GFus}}_{t,1,:} W_Q, \\
K_2 &= \textbf{X}^{\textsc{GFus}}_{t,2,:} W_K, \\
V_2 &= \textbf{X}^{\textsc{GFus}}_{t,2,:} W_V.
\end{aligned}
\label{eqn_1}
\end{equation}
where $W_Q$, $W_K$, $W_V$ are trainable projection matrices that transform the input into new feature subspaces. To compute inter-channel feature correlations, we employ the scaled dot-product attention mechanism, which is formulated as follows:
\begin{equation}
\label{eqn_2}
H_{12} = \text{softmax} \left( \frac{Q_{1} K_{2}^T}{\sqrt{d_k}} \right) V_{2}
\end{equation}
where $Q_{1} K_{2}^T$ computes the similarity of the dot product to measure the feature alignment between two channels. The factor $\frac{1}{\sqrt{d_k}}$ serves as a scaling term to avoid excessively large dot-product values. The following softmax operation redistributes attention weights to amplify critical signals and attenuate noise. By applying attention weights to the value matrix $V_2$, the model synthesizes the output as a channel-aware information combination of its components. Finally, we simultaneously assign the feature interaction results to both involved channels, which is described as follows:
\begin{align}
\label{eqn_3}
{\nu}_1=concat(H_{12},H_{13}), \\
{\nu}_2=concat(H_{12},H_{23}), \\
{\nu}_3=concat(H_{13},H_{23})
\end{align}
where ${\nu}_c$ represents the concatenated features obtained by performing attention fusion between the $c$-th channel and the other channels. The computation of $H_{13}$ and $H_{23}$ follows the same method as $H_{12}$, with the only difference being the replacement of the input channel data. Finally, to dynamically calibrate channel contributions in the fused representation, we implement adaptive per-channel weighting parameters, computed via:
\begin{equation}
\label{eqn_4}
\mathcal{F}_{\text{Glo}} = GAP\left(\sum_{c=1}^{3} \alpha_c \nu_c\right)
\end{equation}
where $\alpha_c$ is trainable weighting parameters that reflect the relative importance of different channels. $GAP(\cdot)$ is the global average pooling operation, which decreases the parameter count and mitigates overfitting risks. In general, our MHA module enables cross-channel information interaction and feature integration, while the trainable weighting parameters further refine the fusion strategy.

\begin{figure}[t]
\centering
\includegraphics[width=0.38\textwidth]{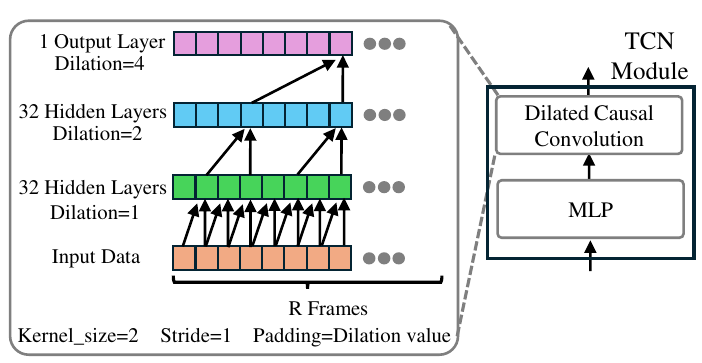}
\caption{Structure of our TCN-based feature fusion module.}
\label{fig:tcn}
\end{figure}

\textbf{Local feature extraction.}
We hereby segment each sample into smaller frames to extract information at a finer resolution, to capture local anomaly features of driving behaviors. Specifically, each sample is divided into 1-second frames, with a sliding step size of 0.5 seconds. As illustrated in Fig.~\ref{fig:architecture}, the process of extracting local features closely mirrors that of global versions, with the key difference being how each frame is processed. During local feature extraction, multi-channel data from a frame are concatenated and then fed to the temporal convolutional network (TCN)~\cite{mnsupervised,cao2024secure}. As depicted in Fig.~\ref{fig:tcn}, the TCN architecture employs causal convolutions (to enforce temporal causality by masking future information) and dilated convolutions (to exponentially expand receptive fields without increasing kernel size). This dual mechanism enables multi-scale temporal pattern extraction while strictly preserving chronological order, a pivotal design for isolating localized anomalies in time-series data. Within the TCN architecture, all input frames are processed in parallel through stacked 1D convolutional layers to model multi-scale temporal dependencies. After passing through this module, the local nonlinear features of multiple frames of a single sample are output, denoted as $\{\textbf{X}_{t,:,:}^{\textsc{LFus},1},\textbf{X}_{t,:,:}^{\textsc{LFus},2},...,\textbf{X}_{t,:,:}^{\textsc{LFus},R}\}$. $R$ denote the total number of segmented temporal frames. During the MHA based fusion stage, all $R$ frame-wise feature matrices are adaptively paired through learnable query-key associations. The final output is the fused feature vector $\mathcal{F}_{\text{Loc}}$ with a size of $1 \times 64$.

\subsection{Behavior Detection Module}
\label{subsec:classfication}
In the driving behavior detection module, the input vector is derived by element-wise addition of $\mathcal{F}_{\text{Glo}}$ and $\mathcal{F}_{\text{Loc}}$. The detection module comprises two fully connected layers, with the hidden and output layers having dimensions of 16 and 2, respectively. Finally, the model performs binary classification based on the output probabilities.

\section{Implementation}
\label{sec:imple}
In this section, we provide descriptions of the basic experimental setup, including data collection, detection model training, baseline models, and evaluation metrics.

\subsection{Data Collection}
\label{subsec:dataCol}
\begin{figure}[t]
\centering
\includegraphics[width=0.34\textwidth]{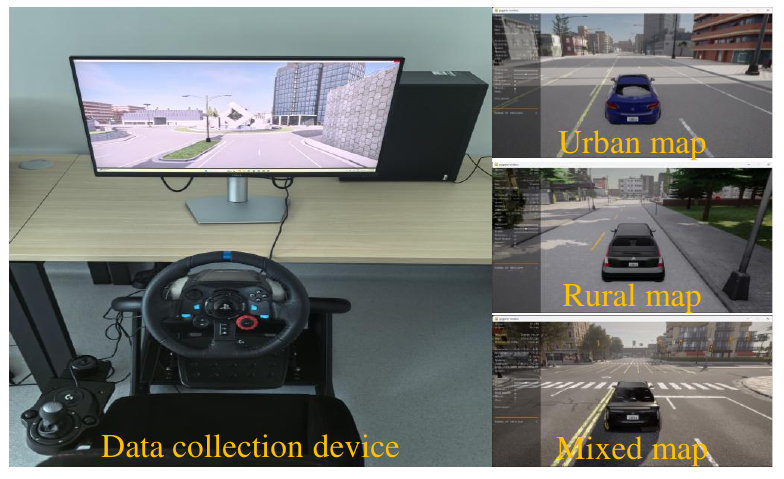}
\caption{Illustration of our data collection environment and three driving maps.}
\label{fig:expEnv}
\end{figure}

Our data collection takes place in a controlled laboratory setting, integrating the Logitech G29 driving platform with the CARLA simulation environment to ensure high-fidelity data acquisition. CARLA, an open-source autonomous driving simulation platform with high-precision capabilities, enhancing both experimental repeatability and scalability. The integration of the G29 driving platform further improves realism by enabling naturalistic driver interactions. As illustrated in Fig.~\ref{fig:expEnv}, we select three maps that encompass both urban, rural, and mixed driving scenarios to capture a diverse range of road conditions. These maps include straight roads, sharp turns, intersections, and slopes, effectively replicating real-world driving complexity. This diversity ensures the creation of a comprehensive dataset that reflects various driving environments. During the experiment, drivers follow pre-established routes and execute essential driving maneuvers, such as straight-line driving and U-turns. Key control inputs, including steering wheel angle, throttle position, and brake pedal travel, are recorded to ensure precise tracking of driving behaviors. To simulate Parkinsonian impairments in driving, three hallmark symptoms (i.e., tremor, bradykinesia, and rigidity) as described in Section~\ref{subsc:sympotm} are integrated by modifying the control inputs to exhibit both continuous and sudden fluctuations. According to existing research~\cite{UPDRS,bloemparkinson,kuosmanen2020let,zhang2022simple}, in straight-line driving scenarios, the frequency of continuous fluctuations ranges from 4 to 6 Hz, with amplitude variations adjusted based on the range observed in healthy driving data. In non-straight-line driving scenarios, the primary impact of motor symptoms on driving control is instability, which can lead to sudden fluctuations. To replicate these sudden fluctuations~\cite{brock2022driving}, we introduce various types of noise (e.g., Spike noise~\cite{nenadic2004spike} and Brownian noise~\cite{zamanzadeh2024deep}) with different amplitudes and durations into the normal data, thereby providing a realistic simulation of driving abnormalities. Ultimately, our dataset comprises 7822 healthy and 6351 abnormal driving samples. All data analysis and processing work is carried out on a desktop equipped with a 13th-generation Intel(R) Core(TM) i7-13620H processor, an NVIDIA(R) GeForce(R) RTX 4060 graphics card, and 16GB LPDDR5 memory.

\subsection{Detection Model Training}
\label{subsec:detMol}
The dataset is partitioned into training and test sets using a default split ratio of 85\%:15\% to evaluate the model's generalization capability. During the training phase, we use the Adam optimizer with an initial learning rate of 0.0001. Additionally, we integrate a learning rate scheduling mechanism that reduces the learning rate to 0.1 times its original value at the seventieth epoch, balancing rapid convergence with training stability. The cross-entropy loss function is adopted due to its inherent suitability for binary classification objectives. Training is conducted over 100 epochs with a batch size of 32. To strengthen generalization performance, a Dropout layer is strategically integrated into the network architecture, effectively diminishing dependence on individual training samples while enhancing model robustness. Systematic hyperparameter optimization via grid search ensures the identification of optimal training configurations.

\subsection{Baseline Models}
\label{subsec:detMol}
Through a thorough review of the literature, we find no prior research that directly corresponds to the driving-specific application of our proposed detection mechanism. Consequently, we select three existing studies~\cite{3300313,9995237,3659627} that focus on detecting PD-related abnormal behaviors within home environments. These methods rely on wearable devices to collect data, leveraging smartphone usage patterns and daily gait activity to distinguish between healthy and abnormal behaviors. To establish a comparative baseline, we replicate the detection models employed in these studies, including support vector machines (SVM), random forests (RF), AdaBoost, and a CNN-Transformer-based neural network (CT), then adapt them to our dataset to ensure compatibility. Finally, to assess the effectiveness of our proposed model, we compare \sname’s accuracy against these existing approaches under identical experimental conditions.

\subsection{Evaluation Metrics}
\label{subsec:metrics}
Identification of normal and pathological abnormal driving behaviors constitutes a binary classification task, where normal driving behavior is assigned a label of 0, while pathological abnormal anomaly is labeled 1. In this study, we leverage \textit{accuracy} as the primary evaluation metric to assess the performance of the model, as it quantifies the proportion of correctly classified samples. To further evaluate the effectiveness of the model in distinguishing between the two classes, we use a confusion matrix to provide a detailed representation of the classification results. The confusion matrix presents the distribution of true and predicted labels in both categories, offering a comprehensive insight into the accuracy of the model's classification.

\begin{figure}[b] %
\centering
\subfigure[1st map]{
\begin{minipage}[t]{0.29\linewidth}
\centering
\includegraphics[width=1\textwidth]{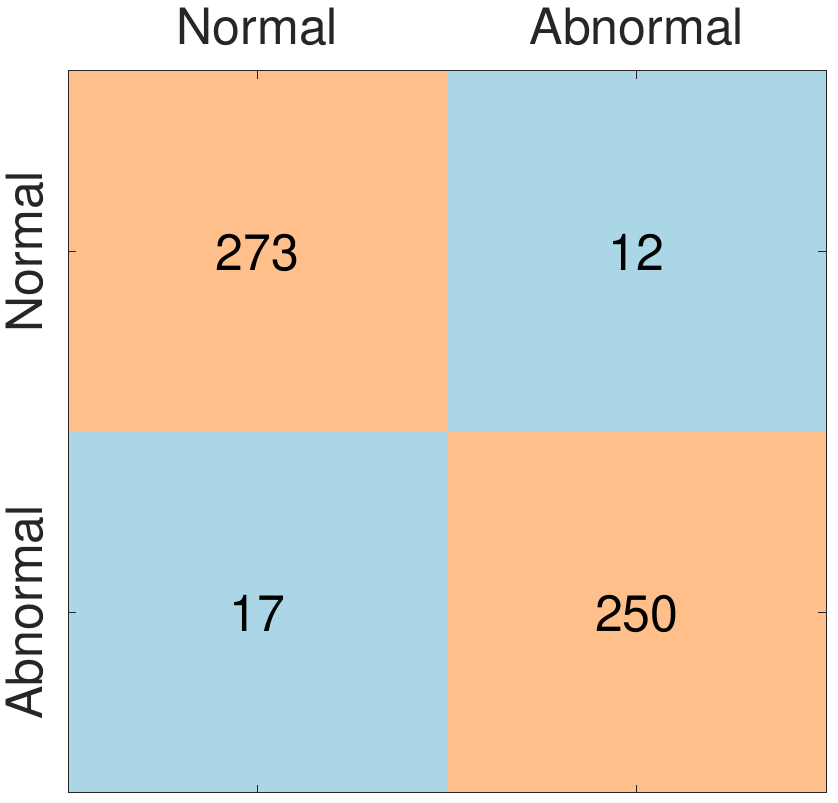}
\label{fig:line}
\end{minipage}
}
\subfigure[2nd map]{
\begin{minipage}[t]{0.29\linewidth}
\centering
\includegraphics[width=1\textwidth]{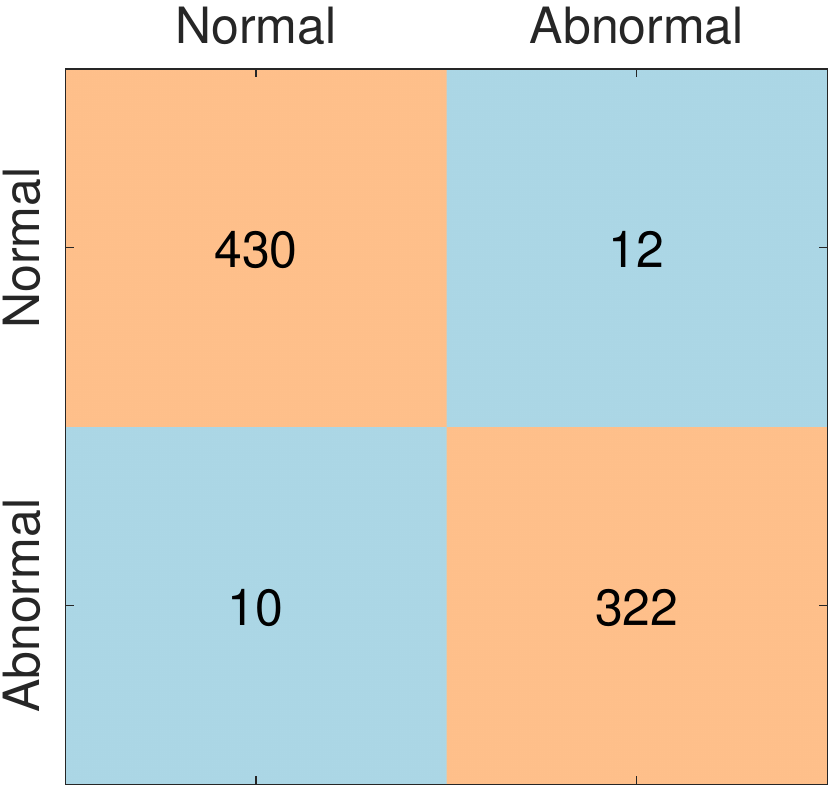}
\label{fig:nonline}
\end{minipage}
}
\subfigure[3rd map]{
\begin{minipage}[t]{0.29\linewidth}
\centering
\includegraphics[width=1\textwidth]{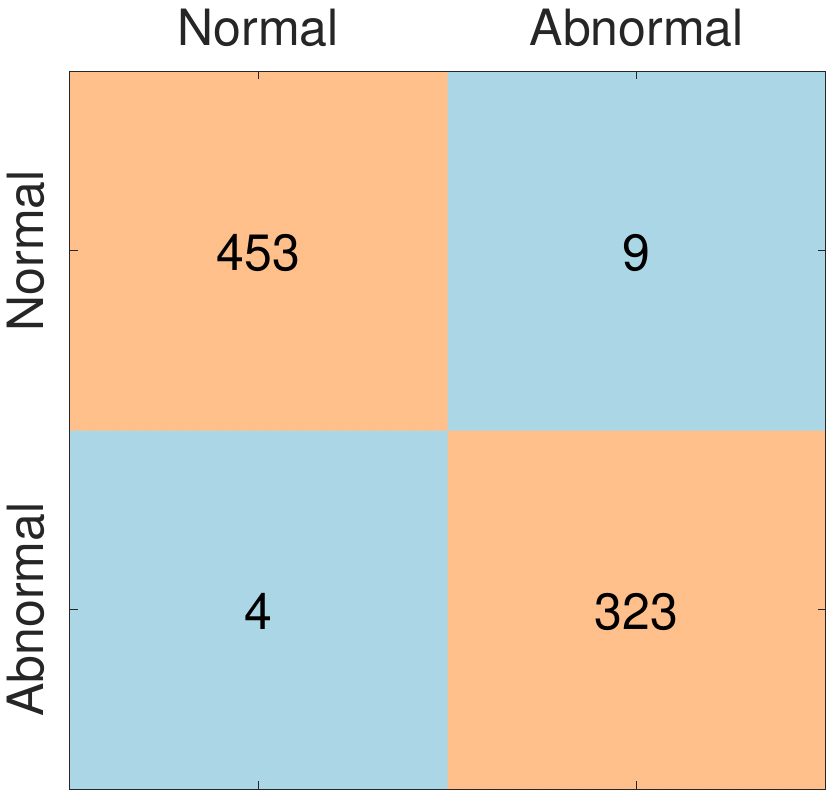}
\label{fig:nonline}
\end{minipage}
}
\caption{Overall detection accuracy in three driving maps.}
\label{fig:overallP}
\end{figure}

\section{Experiment Results}
\label{sec:evaluation}
In this section, we evaluate the performance of \name under different settings and compare it with existing works.

\textbf{Overall performance.}
This experiment aims to assess the overall performance of our framework in detecting abnormal driving behaviors. \name classifies healthy and abnormal driving behaviors across three different map environments to evaluate the system's detection capabilities. As illustrated in Fig.~\ref{fig:overallP}, we count the classification accuracy for each map and give the corresponding confusion matrices. The system achieves an average classification accuracy of 96.8\% across all maps, with only 3.6\% variation between environments. These results indicate that the system effectively distinguishes two-typed drivers, showing a low misclassification rate and strong discriminative power. Furthermore, the minimal fluctuation of accuracy in different environments suggests the robustness and adaptability of the system under varying driving conditions. In summary, our findings validate the feasibility of the detection of PD-related abnormal driving behaviors, demonstrating both high classification accuracy and environmental stability. This study provides empirical support for the integration of driving-based health monitoring systems and offers a novel approach to PD early detection in driving.

\begin{figure}[t]
\subfigure[1st map]{
\begin{minipage}[t]{0.29\linewidth}
\centering
\includegraphics[width=1\textwidth]{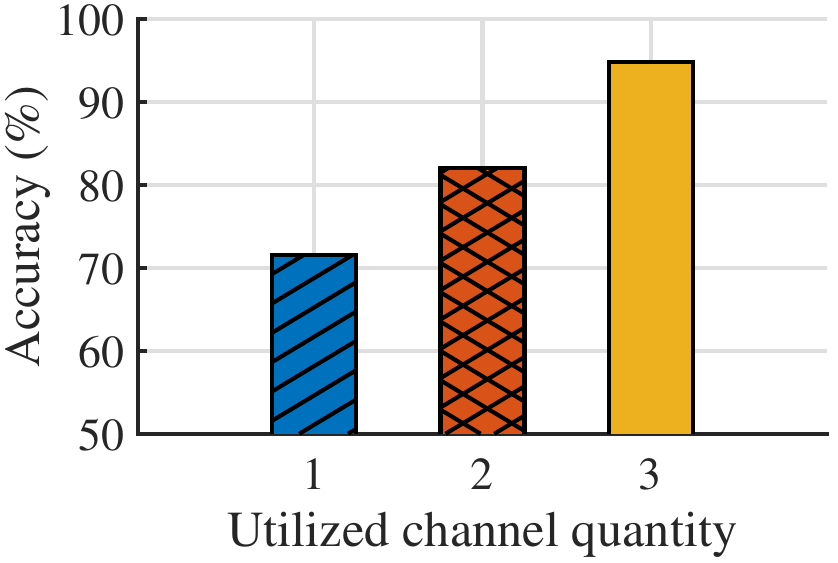}
\label{fig:channel1}
\end{minipage}
}
\subfigure[2nd map]{
\begin{minipage}[t]{0.29\linewidth}
\centering
\includegraphics[width=1\textwidth]{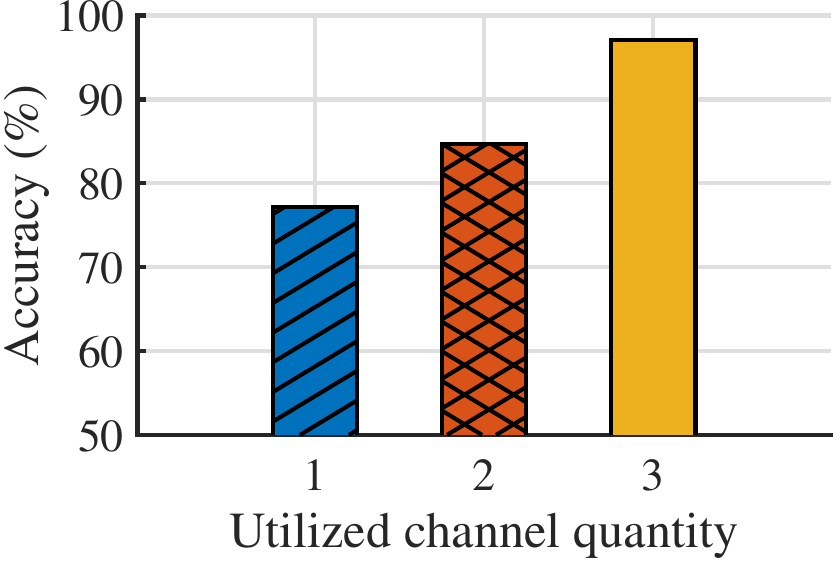}
\label{fig:channel2}
\end{minipage}
}
\subfigure[3rd map]{
\begin{minipage}[t]{0.29\linewidth}
\centering
\includegraphics[width=1\textwidth]{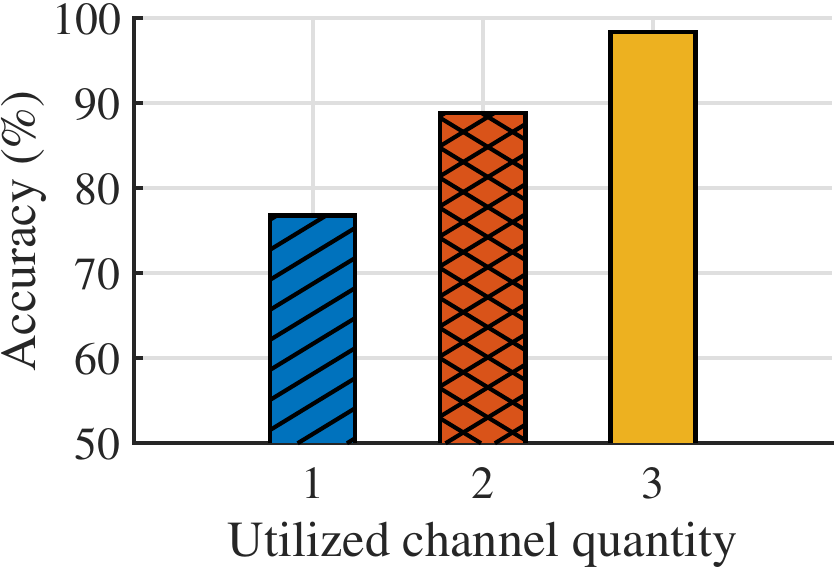}
\label{fig:channel3}
\end{minipage}
}
\caption{Driving behavioral detection performance when using different numbers of channel data in three maps.}
\label{fig:channelP}
\end{figure}

\textbf{The impact of sensing channel count.}
\name utilizes sensing data from three independent driving components to comprehensively characterize driving behaviors, thereby enhancing the classification capability for distinguishing between healthy and PD drivers. The primary objective of this experiment is to assess whether multi-channel information fusion yields information gain and contributes to improve detection performance. To this end, we modify the input configuration of the model, using 1-channel, 2-channel, and 3-channel data for model training and evaluation. This approach enables a quantitative analysis of how varying input modalities impact classification accuracy. As depicted in Fig.~\ref{fig:channelP}, the experimental results show a progressive improvement in classification accuracy as the number of channels increases from 1 to 3, with a 21.6\% increase in the three driving maps. This trend underscores the effectiveness of multi-channel data fusion in augmenting the model’s capacity to capture driving behavioral features, thereby facilitating more precise differentiation between healthy and PD driver driving patterns. Within the context of our study, using multi-channel data for behavioral representation emerges as a critical strategy for improving the performance of abnormal driving pattern detection.

\begin{figure}[b] 
\subfigure[1st map]{
\begin{minipage}[t]{0.29\linewidth}
\centering
\includegraphics[width=1\textwidth]{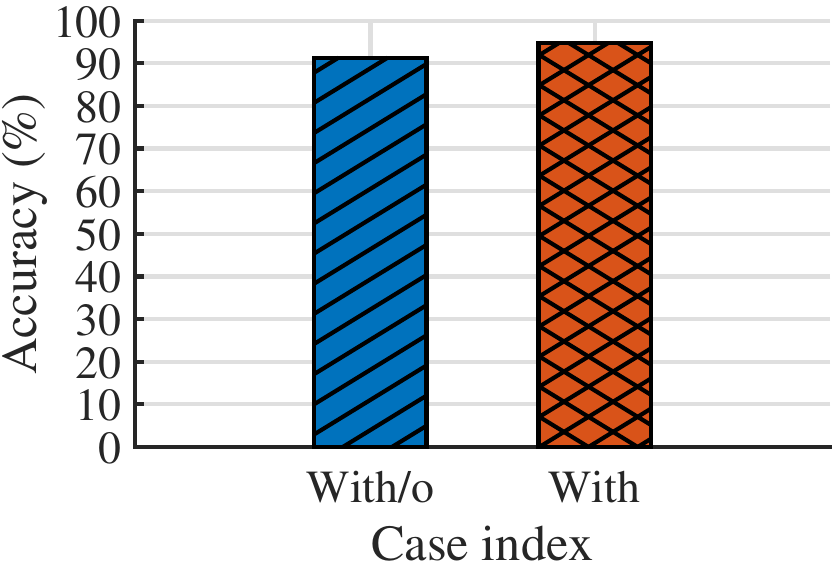}
\label{fig:attn1}
\end{minipage}
}
\subfigure[2nd map]{
\begin{minipage}[t]{0.29\linewidth}
\centering
\includegraphics[width=1\textwidth]{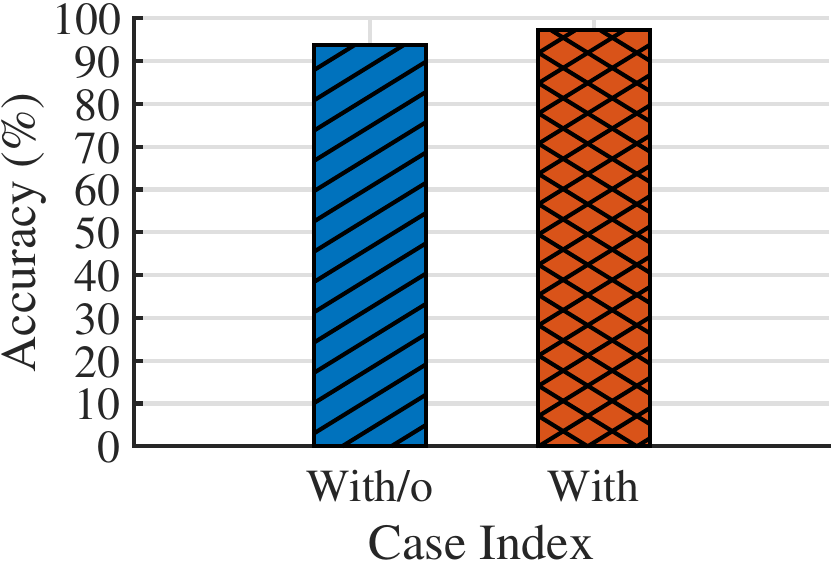}
\label{fig:attn2}
\end{minipage}
}
\subfigure[3rd map]{
\begin{minipage}[t]{0.29\linewidth}
\centering
\includegraphics[width=1\textwidth]{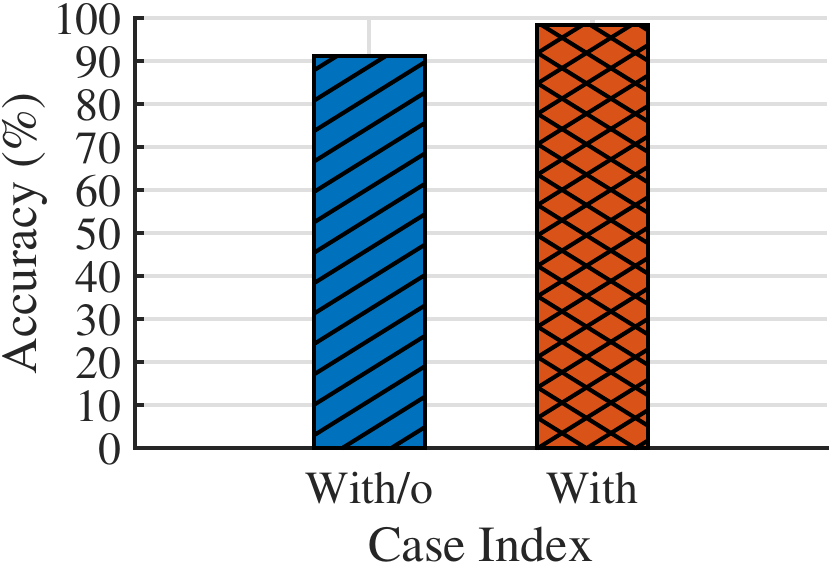}
\label{fig:attn3}
\end{minipage}
}
\caption{Driving behavioral detection performance without and with the attention module in three maps.}
\label{fig:AttP}
\end{figure}

\textbf{The impact of attention module.}
As described in Section~\ref{subsec:feature}, this study leverages an attention mechanism to integrate sensing data across different time frames and channels, enabling a unified representation of driving behaviors. To evaluate the effectiveness of the attention module in our detection framework, we design a comparative experiment in which the model is trained and evaluated under two conditions: one with the attention mechanism enabled and the other with fixed data weights. This approach allows us to quantify its impact on classification performance. As illustrated in Fig.~\ref{fig:AttP}, when the attention mechanism is not utilized, the model's classification accuracy decreased by an average of 4.7\%, indicating a diminished ability to integrate information. This result further validates the significance of the attention mechanism, as it dynamically adjusts the weights of different input features, enabling the model to more accurately capture key driving behavioral patterns and improve classification outcomes. Therefore, using an attention mechanism in the fusion of multi-channel data across time frames is an effective strategy.

\begin{figure}[b]
\centering
\begin{minipage}[t]{0.45\linewidth}
\centering
\includegraphics[width=1\textwidth]{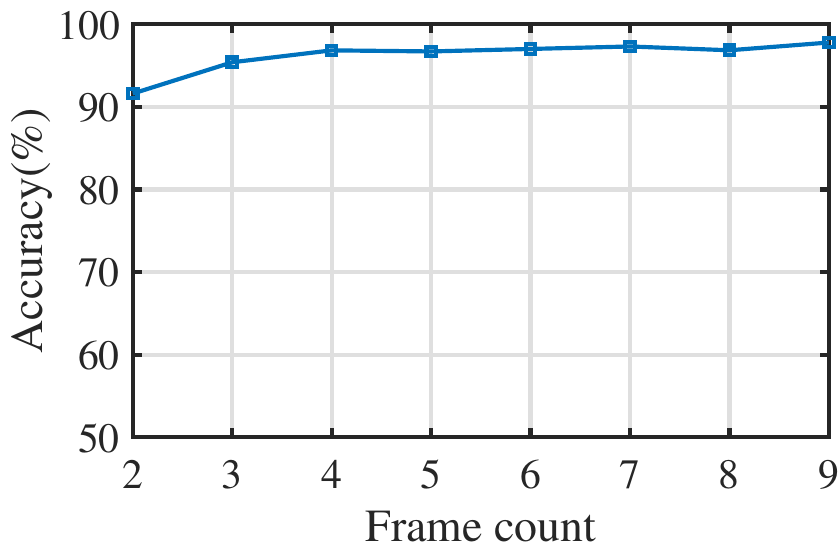}
\caption{Detection performance as varying the data frame count.}
\label{fig:frameP}
\end{minipage}
\hspace{0.1em}
\begin{minipage}[t]{0.45\linewidth}
\centering
\includegraphics[width=1\textwidth]{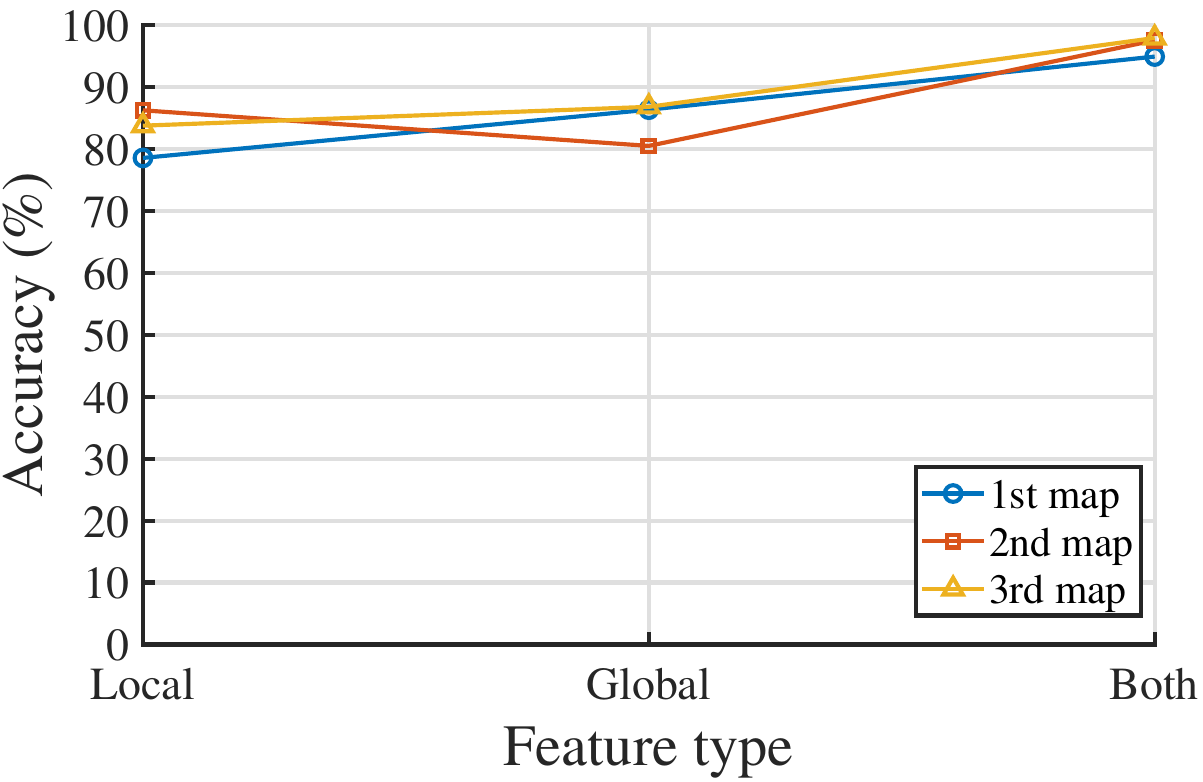}
\caption{Detection performance as using distinct feature types.}
\label{fig:featureP}
\end{minipage}
\end{figure}

\textbf{The impact of data frame count.}
In our study, driving data is segmented into independent time frames, which serve as the fundamental input units for the model. Since the detection of PD behaviors depends on temporal dependencies, it is crucial to incorporate multiple consecutive time frames to effectively capture the dynamic nature of driving patterns. The number of input time frames directly impacts the amount of available information, which may influence detection performance to varying degrees. To systematically investigate the effect of time frame count on classification accuracy, we conduct experiments across three maps, varying the number of input frames and analyzing the resulting accuracy trends. As shown in Fig.~\ref{fig:frameP}, the average accuracy stabilizes when the number of time frames reaches 4, beyond which further increases result in only minimal performance variations. Therefore, in \sname, we set the default number of input frames to 4, striking a balance between detection performance and computational efficiency, ultimately optimizing the model’s overall effectiveness and practicality.

\textbf{The impact of feature types.}
Each sample consists of continuous time frames of 1 second, designed to comprehensively capture driving behavioral characteristics. At a temporal scale, an individual time frame provides fine-grained local driving information, such as instantaneous steering wheel angle and accelerator pedal variations. In contrast, the aggregation of multiple consecutive frames facilitates the extraction of global temporal patterns, such as the execution of a complete right turn. To effectively leverage both micro-scale and macro-scale driving information, our proposed detection model employs an architecture that integrates local and global feature representations, thereby establishing a coherent link between fine-grained driving behaviors and overarching driving trends to enhance classification performance. To systematically assess the efficacy of this feature fusion strategy, we conduct comparative experiments under three conditions: i) utilizing only local features, ii) utilizing only global features, and iii) incorporating both feature types simultaneously. As illustrated in Fig.~\ref{fig:featureP}, experimental results demonstrate that the integration of two-typed features improves average classification accuracy by 22.4\% compared to models relying solely on either feature type. These findings underscore the complementary nature of two-typed features in driving behavioral analysis. Consequently, our proposed feature fusion strategy is instrumental in enhancing the accuracy of PD-related driving behavioral detection.

\begin{figure}[t]
\centering
\begin{minipage}[t]{0.47\linewidth}
\centering
\includegraphics[width=1\textwidth]{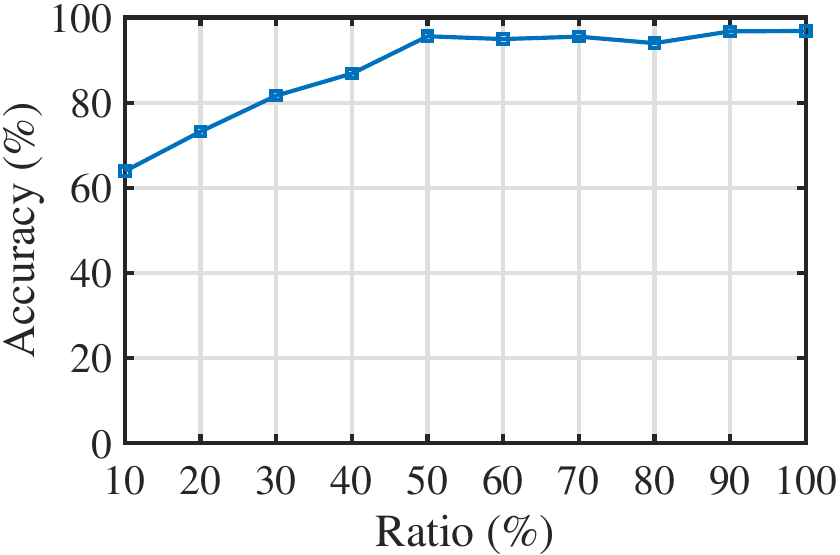}
\caption{Detection performance under different settings of dataset sizes.}
\label{fig:datasizeP}
\end{minipage}
\hspace{0.1em}
\begin{minipage}[t]{0.47\linewidth}
\centering
\includegraphics[width=1\textwidth]{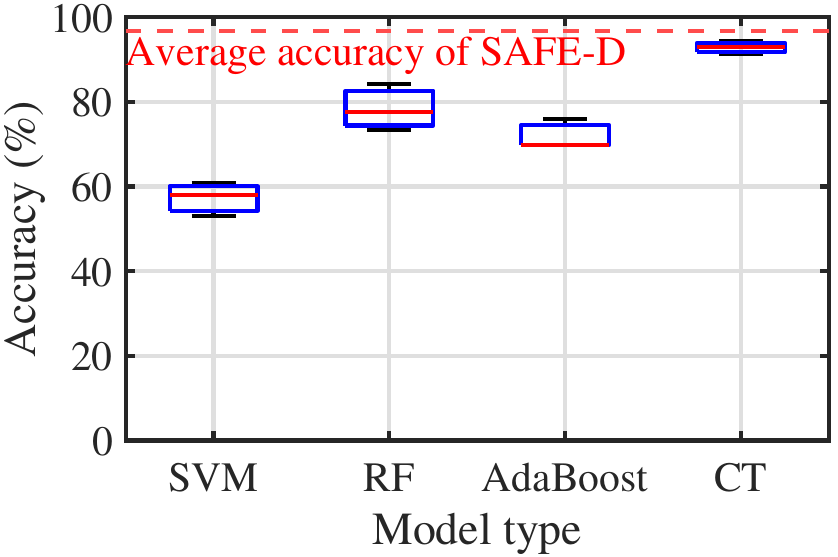}
\caption{Detection performance comparison between \name and existing works.}
\label{fig:compP}
\end{minipage}
\end{figure}

\textbf{The impact of training dataset size.}
The size of the training dataset plays a pivotal role in the development of detection models, as it directly influences the optimization of model parameters and, consequently, its performance. To empirically examine the impact of dataset size, this study incrementally adjusts the ratio, implementing ten different ratio configurations (ranging from 10\% to 100\%, with a 10\% step size) to quantitatively explore the relationship between dataset size and model efficacy. The experimental results (showing in Fig.~\ref{fig:datasizeP}) demonstrate that, in the low data volume phase (training set ratio less than 50\%), model accuracy significantly increased ($+$31.7\%), thereby validating the positive contribution of data augmentation to feature learning. However, when the training set ratio surpasses a threshold of 50\%, model performance is plateaued, with accuracy fluctuations stabilizing within $\pm$3.9\%. This suggests that, beyond this point, the model’s performance improvements are diminishing, and the effect of additional training samples progressively weakens. These findings provide essential theoretical insight into the optimization of behavioral detection models, emphasizing the need for careful consideration in selecting the appropriate data size during training. Consequently, in our context, utilizing 100\% of the available data is sufficient to effectively train the current model, as its performance has already reached stability.

\textbf{Performance comparison between \name and existing works.}
In addition to analyzing the impact of our proposed model’s parameter configurations and submodules, it is crucial to benchmark its performance against existing state-of-the-art methods to comprehensively evaluate its effectiveness. To this end, we conduct a comparative analysis between the \name model and four widely used baseline approaches~\cite{3300313,9995237,3659627}, namely SVM, RF, AdaBoost, and CT. All models are trained and tested under identical experimental settings, utilizing the same dataset split and evaluation metrics, to ensure a fair and consistent comparison. Specifically, detection accuracy is employed as the primary metric for performance evaluation. The experimental results, presented in Fig.~\ref{fig:compP}, reveal that \name consistently outperforms the baseline models in three driving scenarios. \name achieves an average accuracy improvement of 21.4\% across four models and three maps, demonstrating its superior capability in accurately detecting abnormal driving behaviors. Among these models, the performance of CT is the closest to our SAFE-D, as it utilizes the stable and advanced convolutional neural network and Transformer architecture. Nevertheless, our model still maintains an accuracy advantage of 2.6\%. These results highlight the advantages of our proposed multi-channel fusion framework and attention mechanisms in capturing subtle behavioral patterns associated with PD-related driving anomalies. Furthermore, the consistent performance gains across multiple test conditions emphasize the generalization ability of \sname, reinforcing its practical potential for deployment.

\section{Discussions and Future Works}
\label{sec:discussion}

\textit{Integrating multiple pathological indicators for a comprehensive assessment.}
This study explores the potential of using sensing data from in-vehicle components to detect PD-related driving behaviors. It is important to note that abnormal behaviors are not exclusive to PD; other neurological conditions, such as essential tremor and cerebrovascular diseases, may present similar symptoms. As such, no single diagnostic modality can fully capture the pathological state of a PD driver or its impact on driving behaviors. Our proposed mechanism aims to provide users and clinicians with an additional layer of pathological monitoring through risk alerts, not as a standalone clinical diagnostic tool. Clinicians are encouraged to complement these alerts with additional physiological indicators, imaging studies, and cognitive assessments for a more comprehensive evaluation, minimizing misdiagnosis. Furthermore, future advanced driver-assistance systems should account for abnormal behaviors from pathological drivers and mitigate these effects through intelligent control systems.

\textit{Real-world driving data collection and usage.}
Our study, which utilizes laboratory data generated by the Logitech G29 steering wheel and the CARLA simulation platform, examines the potential of analyzing abnormal driving behavior patterns induced by PD. The primary limitation in utilizing real-world data stems from ongoing discussions with medical institutions regarding the methodologies for data usage, with a particular emphasis on safeguarding user privacy and ensuring traffic safety. Given that real-world driving data inherently involves sensitive personal information, it is crucial to adhere to stringent privacy protection regulations. Future research will aim to further strengthen collaborations with medical institutions to identify and implement effective strategies to collect and use real-world driving data, while maintaining robust data security and traffic safety protocols.

\textit{Monitoring trends in individual disease progression.}
\name employs driving behavioral data to preliminarily assess the likelihood of a driver having PD. Future research aims to build on this work by analyzing driving behaviors from confirmed PD to dynamically evaluate disease progression. Central to this approach is the continuous monitoring and analysis of patients’ behavioral patterns across various time instances and driving scenarios, enabling precise tracking of disease evolution. To achieve these objectives, future efforts will combine advanced statistical modeling with deep learning techniques to detect subtle anomalies in driving behaviors and develop personalized models for each patient. Real-time data acquisition and analysis support continuous monitoring of driving statuses, providing timely feedback on disease progression to guide clinical interventions. Overall, this research is expected to not only promote early intervention strategies but also improve patients’ quality of life.

\section{Conclusion}
\label{sec:conclusion}
Our study has systematically studied the impact of PD on driving behaviors in dynamic environments and developed a comprehensive detection framework that focuses on recognizing abnormal driving patterns. We have comprehensively analyzed the manifestations and dynamic variations of motor symptoms of PD in various driving scenarios, incorporating data from three different types of sensors in vehicles, and utilizing attention mechanisms to improve the model's sensitivity to abnormal behaviors. The experimental results have indicated that our proposed framework achieves high accuracy in the detection of abnormal driving behaviors, demonstrating significant potential for practical applications.

\bibliographystyle{IEEEtran}
\bibliography{reference}

\begin{thebibliography}{10}
\providecommand{\url}[1]{#1}
\csname url@samestyle\endcsname
\providecommand{\newblock}{\relax}
\providecommand{\bibinfo}[2]{#2}
\providecommand{\BIBentrySTDinterwordspacing}{\spaceskip=0pt\relax}
\providecommand{\BIBentryALTinterwordstretchfactor}{4}
\providecommand{\BIBentryALTinterwordspacing}{\spaceskip=\fontdimen2\font plus
\BIBentryALTinterwordstretchfactor\fontdimen3\font minus \fontdimen4\font\relax}
\providecommand{\BIBforeignlanguage}[2]{{%
\expandafter\ifx\csname l@#1\endcsname\relax
\typeout{** WARNING: IEEEtran.bst: No hyphenation pattern has been}%
\typeout{** loaded for the language `#1'. Using the pattern for}%
\typeout{** the default language instead.}%
\else
\language=\csname l@#1\endcsname
\fi
#2}}
\providecommand{\BIBdecl}{\relax}
\BIBdecl

\bibitem{baccomera}
M.~H. Baccour, F.~Driewer, E.~Kasneci, and W.~Rosenstiel, ``{Camera-based Eye Blink Detection Algorithm for Assessing Driver Drowsiness},'' in \emph{IEEE Intelligent Vehicles Symposium}, 2019, pp. 987--993.

\bibitem{shahneous}
M.~Shahbakhti, M.~Beiramvand, I.~Rejer, P.~Augustyniak, and et~al, ``{Simultaneous Eye Blink Characterization and Elimination from Low-channel Prefrontal EEG Signals Enhances Driver Drowsiness Detection},'' \emph{IEEE Journal of Biomedical and Health Informatics}, vol.~26, no.~3, pp. 1001--1012, 2021.

\bibitem{yanigueview}
C.~Yang, Z.~Yang, W.~Li, and J.~See, ``Fatigueview: A multi-camera video dataset for vision-based drowsiness detection,'' \emph{IEEE Transactions on Intelligent Transportation Systems}, vol.~24, no.~1, pp. 233--246, 2022.

\bibitem{subasi2022eeg}
A.~Subasi, A.~Saikia, K.~Bagedo, A.~Singh, and A.~Hazarika, ``Eeg-based driver fatigue detection using fawt and multiboosting approaches,'' \emph{IEEE Transactions on Industrial Informatics}, vol.~18, no.~10, pp. 6602--6609, 2022.

\bibitem{xie2020real}
Y.~Xie, F.~Li, Y.~Wu, S.~Yang, and Y.~Wang, ``{Real-time Detection for Drowsy Driving via Acoustic Sensing on Smartphones},'' \emph{IEEE Transactions on Mobile Computing}, vol.~20, no.~8, pp. 2671--2685, 2020.

\bibitem{bloemparkinson}
B.~R. Bloem, M.~S. Okun, and C.~Klein, ``{Parkinson's Disease},'' \emph{The Lancet}, vol. 397, no. 10291, pp. 2284--2303, 2021.

\bibitem{crizzkinson}
A.~M. Crizzle, S.~Classen, and E.~Y. Uc, ``{Parkinson Disease and Driving: An Evidence-based Review},'' \emph{Neurology}, vol.~79, no.~20, pp. 2067--2074, 2012.

\bibitem{milekovspinal}
T.~Milekovic, E.~M. Moraud, N.~Macellari, C.~Moerman, and et~al, ``{A Spinal Cord Neuroprosthesis for Locomotor Deficits due to Parkinson’s Disease},'' \emph{Nature Medicine}, vol.~29, no.~11, pp. 2854--2865, 2023.

\bibitem{aarslankinson}
D.~Aarsland, L.~Batzu, G.~M. Halliday, G.~J. Geurtsen, C.~Ballard, and et~al, ``{Parkinson Disease-associated Cognitive Impairment},'' \emph{Nature reviews Disease primers}, vol.~7, no.~1, p.~47, 2021.

\bibitem{schestysiologic}
P.~Schestatsky, H.~Kumru, J.~Valls-Sole, and et~al, ``{Neurophysiologic Study of Central Pain in Patients with Parkinson Disease},'' \emph{Neurology}, vol.~69, no.~23, pp. 2162--2169, 2007.

\bibitem{kuosmanen2020let}
E.~Kuosmanen, V.~Kan, A.~Visuri, S.~Hosio, and D.~Ferreira, ``{Let's Draw: Detecting and Measuring Parkinson's Disease on Smartphones},'' in \emph{Proceedings of the 2020 CHI Conference on Human Factors in Computing Systems}, Hawaii, USA, 2020, pp. 1--9.

\bibitem{zhang2022simple}
M.~Zhang, N.~Zhao, Y.~Yu, Y.~Zhuang, and et~al, ``{A Simple yet Effective Hand Pose Tremor Classification Algorithm to Diagnosis Parkinsons Disease},'' in \emph{2022 IEEE International Conference on Bioinformatics and Biomedicine}, Florida, USA, 2022, pp. 887--890.

\bibitem{ling2024model}
K.~Ling, H.~Zhao, X.~Fan, X.~Niu, and et~al, ``{Model Touch Pointing and Detect Parkinson's Disease via a Mobile Game},'' \emph{Proceedings of the ACM on Interactive, Mobile, Wearable and Ubiquitous Technologies}, vol.~8, no.~2, pp. 1--24, 2024.

\bibitem{zhang2020pdlens}
H.~Zhang, G.~Guo, C.~Song, C.~Xu, and et~al, ``{PDLens: Smartphone Knows Drug Effectiveness among Parkinson's via Daily-life Activity Fusion},'' in \emph{Proceedings of the 26th Annual International Conference on Mobile Computing and Networking}, Virtual, 2020, pp. 1--14.

\bibitem{jovaltimodal}
F.~Jovan, C.~Morgan, R.~McConville, E.~L. Tonkin, and et~al, ``{Multimodal Indoor Localisation in Parkinson's Disease for Detecting Medication Use: Observational Pilot Study in a Free-Living Setting},'' in \emph{Proceedings of the 29th ACM SIGKDD Conference on Knowledge Discovery and Data Mining}, Washington, USA, 2023, pp. 4273--4283.

\bibitem{yangtificial}
Y.~Yang, Y.~Yuan, G.~Zhang, H.~Wang, and et~al, ``{Artificial Intelligence-enabled Detection and Assessment of Parkinson’s disease Using Nocturnal Breathing Signals},'' \emph{Nature Medicine}, vol.~28, no.~10, pp. 2207--2215, 2022.

\bibitem{anjemark2023car}
L.~Anjemark, H.~Selander, and H.~C. Persson, ``{Car Accidents in Drivers with Parkinson's Disease or Multiple Sclerosis: A Swedish Nationwide Study},'' \emph{European Journal of Neurology}, vol.~30, no.~6, pp. 1631--1638, 2023.

\bibitem{brock2022driving}
P.~Brock, L.~L. Oates, W.~K. Gray, E.~J. Henderson, and et~al, ``{Driving and Parkinson’s Disease: a Survey of the Patient’s Perspective},'' \emph{Journal of Parkinson’s Disease}, vol.~12, no.~1, pp. 465--471, 2022.

\bibitem{gotardi2022parkinson}
G.~C. Gotardi, F.~A. Barbieri, R.~O. Simao, V.~A. Pereira, and et~al, ``{Parkinson’s Disease Affects Gaze Behaviour and Performance of Drivers},'' \emph{Ergonomics}, vol.~65, no.~9, pp. 1302--1311, 2022.

\bibitem{mostafa2019examining}
S.~A. Mostafa, A.~Mustapha, M.~A. Mohammed, R.~I. Hamed, and et~al, ``{Examining Multiple Feature Evaluation and Classification Methods for Improving the Diagnosis of Parkinson’s Disease},'' \emph{Cognitive Systems Research}, vol.~54, pp. 90--99, 2019.

\bibitem{UPDRS}
{International Parkinson and Movement Disorder Society}, ``{MDS-Unified Parkinson's Disease Rating Scale Certificate Program},'' \url{https://mds.movementdisorders.org/updrs}, 2025, online; accessed 10 January 2025.

\bibitem{3652181}
G.-H. Li, H.-C. Chiang, Y.-C. Li, S.~Shirmohammadi, and C.-H. Hsu, ``{A Driver Activity Dataset with Multiple RGB-D Cameras and mmWave Radars},'' in \emph{Proceedings of the 15th ACM Multimedia Systems Conference}, New York, NY, USA, 2024, p. 360–366.

\bibitem{3581335}
A.~Hafizi, J.~Henderson, A.~Neshati, W.~Zhou, E.~Lank, and D.~Vogel, ``{In-vehicle Performance and Distraction for Midair and Touch Directional Gestures},'' in \emph{Proceedings of the 2023 CHI Conference on Human Factors in Computing Systems}, New York, NY, USA, 2023.

\bibitem{10265761}
H.~Yang, H.~Liu, Z.~Hu, A.-T. Nguyen, T.-M. Guerra, and C.~Lv, ``{Quantitative Identification of Driver Distraction: A Weakly Supervised Contrastive Learning Approach},'' \emph{IEEE Transactions on Intelligent Transportation Systems}, vol.~25, no.~2, pp. 2034--2045, 2024.

\bibitem{9507391}
S.~Ansari, F.~Naghdy, H.~Du, and Y.~N. Pahnwar, ``{Driver Mental Fatigue Detection Based on Head Posture Using New Modified reLU-BiLSTM Deep Neural Network},'' \emph{IEEE Transactions on Intelligent Transportation Systems}, vol.~23, no.~8, pp. 10\,957--10\,969, 2022.

\bibitem{10382460}
L.~Yang, H.~Yang, H.~Wei, Z.~Hu, and C.~Lv, ``{Video-Based Driver Drowsiness Detection With Optimised Utilization of Key Facial Features},'' \emph{IEEE Transactions on Intelligent Transportation Systems}, vol.~25, no.~7, pp. 6938--6950, 2024.

\bibitem{9758635}
T.~Huang and R.~Fu, ``{Driver Distraction Detection Based on the True Driver’s Focus of Attention},'' \emph{IEEE Transactions on Intelligent Transportation Systems}, vol.~23, no.~10, pp. 19\,374--19\,386, 2022.

\bibitem{3460938}
M.~Braun, F.~Weber, and F.~Alt, ``{Affective Automotive User Interfaces–Reviewing the State of Driver Affect Research and Emotion Regulation in the Car},'' \emph{ACM Computing Surveys}, vol.~54, no.~7, 2021.

\bibitem{8859275}
L.~Zhang, L.~Yan, Y.~Fang, X.~Fang, and X.~Huang, ``{A Machine Learning-Based Defensive Alerting System Against Reckless Driving in Vehicular Networks},'' \emph{IEEE Transactions on Vehicular Technology}, vol.~68, no.~12, pp. 12\,227--12\,238, 2019.

\bibitem{7478592}
L.~Fridman, P.~Langhans, J.~Lee, and B.~Reimer, ``{Driver Gaze Region Estimation without Use of Eye Movement},'' \emph{IEEE Intelligent Systems}, vol.~31, no.~3, pp. 49--56, 2016.

\bibitem{519267}
K.~Das, M.~Papakostas, K.~Riani, A.~Gasiorowski, M.~Abouelenien, and et~al, ``{Detection and Recognition of Driver Distraction Using Multimodal Signals},'' \emph{ACM Transactions on Interactive Intelligent Systems}, vol.~12, no.~4, 2022.

\bibitem{9507390}
S.~Jha, M.~F. Marzban, T.~Hu, M.~H. Mahmoud, N.~Al-Dhahir, and C.~Busso, ``{The Multimodal Driver Monitoring Database: A Naturalistic Corpus to Study Driver Attention},'' \emph{IEEE Transactions on Intelligent Transportation Systems}, vol.~23, no.~8, pp. 10\,736--10\,752, 2022.

\bibitem{3300313}
F.~Tian, X.~Fan, J.~Fan, Y.~Zhu, and et~al, ``{What Can Gestures Tell? Detecting Motor Impairment in Early Parkinson's from Common Touch Gestural Interactions},'' in \emph{Proceedings of the 2019 CHI Conference on Human Factors in Computing Systems}, San Francisco, USA, 2019, p. 1–14.

\bibitem{9995237}
L.~DArco, H.~Wang, and H.~Zheng, ``{A Rapid Detection of Parkinson’s Disease using Smart Insoles: A Statistical and Machine Learning Approach},'' in \emph{IEEE International Conference on Bioinformatics and Biomedicine}, Vancouver, Canada, 2022, pp. 2985--2992.

\bibitem{3659627}
K.~Ling, H.~Zhao, X.~Fan, X.~Niu, W.~Yin, Y.~Liu, C.~Wang, and X.~Bi, ``{Model Touch Pointing and Detect Parkinson's Disease via a Mobile Game},'' \emph{Proceedings of the ACM on Interactive, Mobile, Wearable and Ubiquitous Technologies}, vol.~8, no.~2, 2024.

\bibitem{azadani2021driving}
M.~N. Azadani and A.~Boukerche, ``{Driving Behavior Analysis Guidelines for Intelligent Transportation Systems},'' \emph{IEEE Transactions on Intelligent Transportation Systems}, vol.~23, no.~7, pp. 6027--6045, 2021.

\bibitem{juncen2023mmdrive}
Z.~Juncen, J.~Cao, Y.~Yang, W.~Ren, and H.~Han, ``{mmDrive: Fine-grained Fatigue Driving Detection Using mmWave Radar},'' \emph{ACM Transactions on Internet of Things}, vol.~4, no.~4, pp. 1--30, 2023.

\bibitem{fang2021spatial}
Z.~Fang, Q.~Long, G.~Song, and K.~Xie, ``{Spatial-temporal Graph ODE Networks for Traffic Flow Forecasting},'' in \emph{Proceedings of the 27th ACM SIGKDD conference on knowledge discovery \& data mining}, Virtual, 2021, pp. 364--373.

\bibitem{wang2021prediction}
F.~Wang, J.~Xu, C.~Liu, R.~Zhou, and P.~Zhao, ``{On Prediction of Traffic Flows in Smart Cities: a Multitask Deep Learning based Approach},'' \emph{World Wide Web}, vol.~24, no.~3, pp. 805--823, 2021.

\bibitem{luo2022estnet}
G.~Luo, H.~Zhang, Q.~Yuan, J.~Li, and F.-Y. Wang, ``{ESTNet: Embedded Spatial-temporal Network for Modeling Traffic Flow Dynamics},'' \emph{IEEE Transactions on Intelligent Transportation Systems}, vol.~23, no.~10, pp. 19\,201--19\,212, 2022.

\bibitem{he2016deep}
K.~He, X.~Zhang, S.~Ren, and J.~Sun, ``{Deep Residual Learning for Image Recognition},'' in \emph{Proceedings of the IEEE Conference on Computer Vision and Pattern Recognition}, 2016, pp. 770--778.

\bibitem{cao2022handkey}
H.~Cao, D.~Liu, H.~Jiang, and et~al, ``{HandKey: Knocking-triggered robust vibration signature for keyless unlocking},'' \emph{IEEE Transactions on Mobile Computing}, vol.~23, no.~1, pp. 520--534, 2022.

\bibitem{zhaoethinking}
B.~Zhao, H.~Xing, X.~Wang, F.~Song, and Z.~Xiao, ``{Rethinking Attention Mechanism in Time Series Classification},'' \emph{Information Sciences}, vol. 627, pp. 97--114, 2023.

\bibitem{mnsupervised}
R.~Mo, Y.~Pei, N.~V. Venkatarayalu, P.~N. Joseph, and et~al, ``{Unsupervised TCN-AE-based Outlier Detection for Time Series with Seasonality and Trend for Cellular Networks},'' \emph{IEEE Transactions on Wireless Communications}, vol.~22, no.~5, pp. 3114--3127, 2022.

\bibitem{cao2024secure}
H.~Cao, L.~Yuan, G.~Xu, Z.~He, Z.~Fang, and Y.~Fang, ``Secure traffic sign recognition: An attention-enabled universal image inpainting mechanism against light patch attacks,'' \emph{arXiv preprint arXiv:2409.04133}, 2024.

\bibitem{nenadic2004spike}
Z.~Nenadic and J.~W. Burdick, ``{Spike Detection Using the Continuous Wavelet Transform},'' \emph{IEEE transactions on Biomedical Engineering}, vol.~52, no.~1, pp. 74--87, 2004.

\bibitem{zamanzadeh2024deep}
Z.~Zamanzadeh~Darban, G.~I. Webb, S.~Pan, C.~Aggarwal, and M.~Salehi, ``{Deep Learning for Time Series Anomaly Detection: A Survey},'' \emph{ACM Computing Surveys}, vol.~57, no.~1, pp. 1--42, 2024.

\end{thebibliography}

\end{document}